\documentclass[letterpaper, 10 pt, conference]{ieeeconf}  

\IEEEoverridecommandlockouts                              
\usepackage{floatrow}
\newfloatcommand{capbtabbox}{table}[][\FBwidth]
\usepackage{graphicx}
\usepackage{comment}
\usepackage{amsmath,amssymb} 
\usepackage{color}
\usepackage{hyperref}
\usepackage{subfigure}
\usepackage{booktabs}

\title{RGB-D SLAM with Structural Regularities}

\author{Yanyan Li$^{1}$, Raza Yunus$^{1}$, Nikolas Brasch$^{1}$, Nassir Navab$^{1,2}$ and Federico Tombari$^{1,3}$
\thanks{$^1$:Technical University of Munich, Germany; {\tt\small \{yanyan.li, raza.yunus, nikolas.brasch, nassir.navab, federico.tombari\}@tum.de}; $^{2}$:Johns Hopkins University, USA;$^3$:Google Inc.}
}

\begin{document}

\maketitle
\thispagestyle{empty}
\pagestyle{empty}


\begin{abstract}
This work proposes a RGB-D SLAM system specifically designed for structured environments and aimed at improved tracking and mapping accuracy by relying on geometric features that are extracted from the surrounding. Structured environments offer, in addition to points, also an abundance of geometrical features such as lines and planes, which we exploit to design both the tracking and mapping components of our SLAM system. For the tracking part, we explore geometric relationships between these features based on the assumption of a Manhattan World (MW). We propose a decoupling-refinement method based on points, lines, and planes, as well as the use of Manhattan relationships in an additional pose refinement module. For the mapping part, different levels of maps from sparse to dense are reconstructed at a low computational cost. We propose an instance-wise meshing strategy to build a dense map by meshing plane instances independently. The overall performance in terms of pose estimation and reconstruction is evaluated on public benchmarks and shows improved performance compared to state-of-the-art methods. The code is released at \url{https://github.com/yanyan-li/PlanarSLAM}. 
\end{abstract}

\section{Introduction}
Visual Simultaneous Localization and Mapping (SLAM) algorithms are used to estimate the 6D camera pose while reconstructing the surrounding unknown environment. They have shown to be useful in a wide range of applications, such as autonomous robots, self-driving cars and augmented/virtual reality, where camera pose estimation enables cars, robots and mobile devices to localize themselves, while the dense map provides a representation of the environment, \textit{e.g.} for robot-environment or human-environment interaction.

Many SLAM applications have to deal with structured scenes, \textit{i.e.} man-made environments that are usually characterized by low-textured surfaces - a typical example is an indoor scene, or an outdoor parking place. This induces a lack of visual features, that visual SLAM systems typically leverage to improve camera pose estimation and/or 3D reconstruction, \textit{e.g.} by carrying out loop closure and bundle adjustment to reduce drift.
In order to deal with structured scenes, specific SLAM methods based on points and line segments, like S-SLAM~\cite{Li2020SSLAM},  Stereo-PLSLAM~\cite{gomezojeda2017pl-slam:}, PLVO~\cite{lu2015robust}, Mono-PLSLAM~\cite{pumarola2017pl-slam:} and Probabilistic-VO \cite{proencca2018probabilistic} have been proposed, extending the working environment to scenes where more lines than points can be detected. SP-SLAM~\cite{zhang2019point} merges plane features into ORB-SLAM2~\cite{murartal2017orb-slam2:}, achieving robust results in low-textured scenes.  

\begin{figure}
    \centering
\includegraphics[scale=0.34]{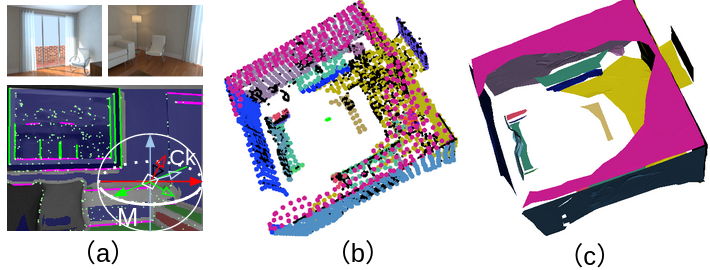}
    \caption{RGB-D SLAM system. (a) Examples of a typical structured scene, and 2D features and orthogonal lines and planes segmentation. (b) Point cloud including points, lines and planes. (c) Real-time mesh on a CPU.}
    \label{fig:t}
\end{figure}
For the reconstruction, there are sparse, semi-dense and dense methods. Compared to the first two classes, which only provide incomplete maps, dense reconstruction is required to provide sufficient information for applications such as robot-environment interaction and 3D scene understanding. Many algorithms have been proposed to reconstruct indoor scenes via RGB-D sensors. KinectFusion~\cite{newcombe2011kinectfusion} is a pioneering work relying on the truncated signed distance field (TSDF) representation of the map. In order to reconstruct large scale scenarios, surfel-based methods, like ElasticFusion~\cite{whelan2015elasticfusion}, were proposed. Instead of reconstructing each pixel, Wang et al.~\cite{wang2019real} extracts superpixels from RGB images and depth maps, which is more efficient but still has redundant information especially in indoor scenarios where large planes can be commonly found.

In this paper, we build on our monocular Structure-SLAM~\cite{Li2020SSLAM} and propose a robust RGB-D SLAM system specifically designed to deal with structured environments, which improves tracking and mapping at the same time. Figure \ref{fig:t} illustrates the components of such structured scenes, which contains points, lines and plane segments.
Following the decoupling strategy of Structure-SLAM, we estimate a drift-free rotation matrix first, and then the 3-DoF translation. The initial rotation and translation are refined via a map-to-frame strategy.
Different to~\cite{Li2020SSLAM,kim2018low,kim2018linear}, plane features are merged into our Manhattan-based framework, which is used to estimation the initial translation vector and retain Manhattan relationships as constrains in the refinement module. Furthermore, an efficient meshing module is proposed that reconstructs the scene structure based on the obtained planar regions in the sparse map. 
In summary our contributions are:
\begin{itemize}
    \item Based on the concept of MW-based decoupled pose estimation, we improve the translation estimation by combining point and line features with planes and an additional pose refinement step with Manhattan relationships.
    \item We propose a planar instance-wise mesh based reconstruction method generating a compact representation of the environment from a sparse point cloud.
    \item A general framework for real-time RGB-D SLAM where these components are used to localize and map under structured environments with high accuracy. 
\end{itemize}
We evaluate the performance of our approach in terms of both camera pose estimation and reconstruction on public benchmarks, showing improved performance compared to state-of-the-art methods.

\section{Related work}
In the following we review the literature related to RGB-D based SLAM systems as well as methods leveraging structural regularity as the MW assumption.
\paragraph{RGB-D SLAM.}
In~\cite{raposo2013plane,taguchi2013point} it was proposed to use planes over point features whenever possible, as the averaging over multiple depth measurements reduces the noise significantly.
In Dense Planar SLAM~\cite{salas2014dense} surfels belonging to the same planar areas are smoothed by fitting a plane to them and back-projecting the surfels onto the plane.
Le \textit{et al.}~\cite{le2017dense} rely on a scene layout consisting of a ground plane and several walls, and use dynamic programming to infer a sequentially consistent assignment of pixels to planes. In Probabilistic-VO~\cite{proencca2018probabilistic}, the uncertainties of points, lines and planes are modelled explicitly and used during pose estimation, where points, lines and planes are represented in a uniform framework in \cite{nardi2019unified}.
A direct SLAM system combining photo-metric and geometric terms is proposed in DVO-SLAM\cite{kerl2013dense} and extended in CPA-SLAM~\cite{ma2016cpa} with global planes, where depth measurements are assigned to the global planes with weights. 

\paragraph{Dense Reconstruction.}
While the aforementioned methods have the goal to estimate precise poses and therefore only maintain a map with the most reliable information, several works have been proposed with the goal to create a complete dense reconstruction of the environment. KinetFusion~\cite{newcombe2011kinectfusion} and ElasticFusion~\cite{whelan2015elasticfusion} explore dense reconstruction for RGB-D sensors. The first method fuses all depth data into a volumetric dense representation, which is used to track the camera pose using ICP. The size of the map is usually limited in volumetric methods due to memory constraints.
Different from KinectFusion, ElasticFusion is a map-centric system that reconstructs surfel-based maps of the environment. In order to decrease the number of surfels in the map, superpixel-based surfels are proposed by~\cite{wang2019real}, which reduces the number of surfels compared with ElasticFusion.
Recently BAD-SLAM~\cite{schops2019bad} proposed a direct bundle-adjustment approach for RGB-D SLAM.
In \cite{schops2019surfelmeshing} a textured mesh is extracted from a dense surfel cloud.
A direct mesh based reconstruction approach for RGB-D sensors was proposed in \cite{schreiberhuber2019scalablefusion}.

\paragraph{Structural Regularity.}
A line of works exploits additional constrains and regularities in the world, to improve the reconstruction performance. In \cite{straub2015real} and \cite{zhou2016divide} the authors showed that the rotation estimation error is the main reason for long-term drift.

A branch-and-bound framework for Manhattan Frame estimation is proposed in~\cite{joo2016globally}. In MVO~\cite{zhou2016divide} a method using mean shift on the unit sphere is used to find the transformation between the MW and the current frame. When only planes are used for the rotation estimation as in OPVO~\cite{kim2017visual} at least 2 orthogonal planes must be detected in each frame, the addition of vanishing points extracted from lines can be used alternatively, as done in LPVO~\cite{kim2018low}. The methods mentioned above use point features to estimate the translation. Structure-SLAM~\cite{Li2020SSLAM} is a monocular system that predicts normals via a convolutional neural network leverages normals with points and lines in a decoupling strategy. Since predicted normals are not as accurate as those computed from a depth map, the system provide a refinement/fallback module based on points and lines. Compared with Structure-SLAM, optimized vanishing points of lines and plane features are used for rotation and translation in this work. Then, the fallback part is removed and the refinement part incorporates geometric relationship of planes. Instead of the sparse point-line map, a dense mesh as output is more useful for robotics applications.
L-SLAM~\cite{kim2018linear} is also based on the MW assumption, which obtains translation, rotation and pixels of potential planar regions from LPVO. Then it refines 3D translational and 1-D plane offsets with a linear Kalman Filter. However, we use a more robust front-end for initial translation estimation. Furthermore, the 6D pose refinement step is used to optimize rotation and translation simultaneously and allows an offset to the initial rotation from MW, which is more robust to non-MW (curved surfaces and few planar regions) compared with L-SLAM and LPVO (see Figure~\ref{fig:lr-k3}). 

\section{Proposed framework} 
\label{sec:structure_analysis}
\begin{figure}
    \centering
    \includegraphics[scale=0.28]{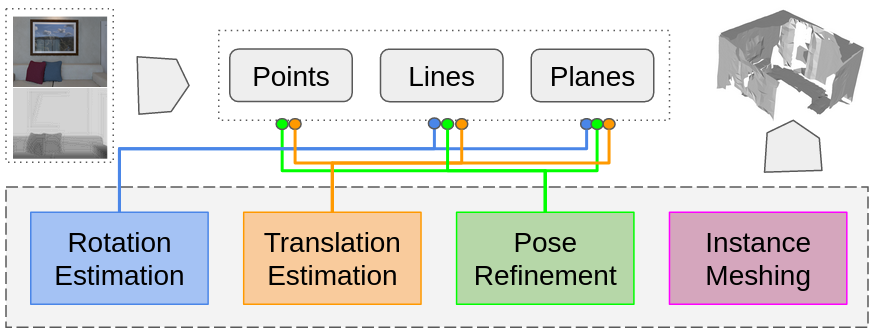}
    \caption{Overview of the proposed framework. Point, line and plane features are extracted from the RGB-D frame. Rotation and translation are estimated in a decoupled fashion first and refined afterwards. The planar segments are used to create a mesh-based reconstruction of the environment.}
    \label{fig:overview}
\end{figure}
Given a sequence of RGB-D frames from a structured environment, the goal of our method is to reconstruct the 3D scene while simultaneously estimating the 6D camera pose at each frame.
Section~\ref{sec:tracking} provides an overview of the proposed tracking pipeline, which decouples rotation and translation, while section~\ref{sec:mapping} describes different types of mapping presentations generated by the system. We now describe the system's underlying features and structural components. 

\subsection{Extended feature set}
In our method we use ORB features~\cite{rublee2011orb}, which are fast to extract and match. In low-textured environments, it is hard to extract sufficient points for robust pose estimation, therefore we extend the feature set with lines, which are extracted using the LSD~\cite{von2010lsd} approach, as described by LBD~\cite{zhang2013efficient}. Furthermore, it is common to find non-textured planar regions in indoor environments, where plane instances extracted from the depth maps are valuable cues to extend points and lines. 
Planes are detected using the connected component analysis method~\cite{trevor2013efficient}. They are represented by the Hessian normal form $\pi = (\hat{n}, d)$, where $\hat{n} = (n_x, n_y, n_z)$ is the normal of the plane, representing its orientation and $d$ is the distance from the camera origin to the plane. 

\paragraph{Points and lines}

After the extraction of 2D point features  $x_j=(u_j,v_j)$ and line segments  $l_j=(x_{j,start},x_{j,end})$ in frame $F_i$, we can back-project points and lines using the camera intrinsic parameters and the depth map to obtain 3D points $X_j$ and 3D lines $L_j$. 
The depth map is not always correct, especially at depth discontinuities \textit{e.g.} object boundaries. Therefore a robust fitting method for 3D lines is needed. First, we count the number of pixels with non-zero depth values intersected by the detected line segment. If the number exceeds a certain threshold, the 3D line $L_j$ will be estimated via RANSAC to remove potential outliers. 
 
\paragraph{Normals and planes.}
Smooth normals are computed by averaging the tangential vectors from the depth image inside a patch of $10\times10$ pixels using integral images. After plane detection, we use the strategy of~\cite{zhang2019point} to associate the observed planes with those present in the map. To match an observed plane with one from the map, we first check the angle between their normals. If it is below the threshold $\theta_n$, we check the point-to-plane distance between them. The plane which has the minimum distance to the observed plane, and also lies below the distance threshold $\theta_P$, is matched to the observed plane. In the experiments, $\theta_n$ and $\theta_P$ are set at 10 degrees and 0.1 m respectively.
Furthermore, we also keep parallel and perpendicular relationships~\cite{zhang2019point} between the map planes to leverage additional constraints during the tracking process. These are determined by the angle between the plane normals. Since they only provide constraints for orientation, we do not consider their distance.

\subsection{Decoupling pose estimation and refinement.}
To reduce error propagation between frames, we build on our monocular 
architecture~\cite{Li2020SSLAM} that computes rotational motion based on the MW assumption. Then the corresponding translational motion is estimated by features, with the fixed rotation computed from last step. 
In the font-end of this work, we use optimized lines for rotation estimation and planes for translation estimation.

Differently to Structure-SLAM~\cite{Li2020SSLAM} that uses a point-line local map to optimize translation and rotation together, we leverage planes in the local map and also make use of the geometric relationship (parallel and perpendicular) of those planes as constrains, which improves the accuracy of the system as it will be shown in Figure~\ref{fig:rpe} and Table~\ref{table:ICL}.


\section{Tracking}
\label{sec:tracking}
Differently from traditional pose estimation methods, we decouple the 6D camera pose into rotation and translation. Based on the MW assumption, we obtain the rotational motion $R_{c_im}$ between the MW and camera $c_i$. In this way, the rotation estimation will not be affected by the pose of the last frame or last keyframe, which reduces drift effectively. Afterwards, point, line and plane features as well as the initial rotation matrix are used for translation estimation, which consists of just 3 Degrees-of-Freedom (DoFs). 

\subsection{Rotation estimation}

Instead of tracking the camera from frame-to-frame directly, the drift-free rotation estimation method estimates the rotation $R_{cm}$ between each frame and the Manhattan coordinate frame, by modeling the indoor environments as a MW, thus reducing the drift generated from frame-to-frame tracking. As shown in Figure~\ref{fig:t}, Manhattan coordinate frames can be aligned to the starting frame of the camera via $R_{k+1,m}$. Generally, the coordinate of the first frame is regarded as the world frame, \textit{i.e.} $R_{1,m}=R_{m,w}^T$. So we can obtain pose in the world coordinate by using,
\begin{equation}
    R_{k+1,w} = R_{k+1, m}R_{m,w}
\end{equation}

Here $R_{m,w}$ represents the relation from the world to MW, which is obtained by the MW initialization step and $R_{k+1, m}$ is the relation from MW to the $(k+1)^th$ frame. These two matrices are computed via a sphere mean-shift method~\cite{zhou2016divide}, where the normals and normalized vanishing directions are projected onto the tangent planes of the current rotation estimate. Then a mean shift step is performed on the tangent planes, which generates new centers and back-projects them to the sphere as new estimates.
We refer the reader to~\cite{zhou2016divide} and~\cite{kim2017visual} for more details on the sphere mean-shift method. 
%
To handle difficult scenes where only one or no plane at all is detected, we feed the unit sphere with both vanishing directions of the refined 3D lines and surface normals of planes, which is a more robust approach than~\cite{kim2017visual,Li2020SSLAM} under these challenging conditions. 

\subsection{Translation estimation}

After estimating the rotation, points, lines and planes are used to estimate the translation. We re-project 3D points from the last frame into the current one and define the error function, based on the re-projection error, as follows,
\begin{equation}
e_{k,j}^p=p_k-\Pi( R_{k,j}P_{j}+t_{k,j})
\label{e_p}
\end{equation}
where $\Pi (\cdot)$ is the projection function. Since the rotation matrix $R_{k,j}$ has been obtained in the last step, we fix the rotation and only estimate the translation using the Jacobian matrix corresponding to~\eqref{e_p}.

As for lines, we obtain the normalized line function from the 2D endpoints $p_{start}$ and $p_{end}$ as follows
\begin{equation}
l=[p_{start}\times p_{end}]/[\|p_{start}\|\|p_{end}\|]=(a,b,c).
\end{equation}

Then, we formulate the error function based on the point-to-line distance~\cite{pumarola2017pl-slam:} between $l$ and the projected 3D endpoints $P_{start}$ and $P_{end}$ from the matched 3D line in the keyframe. For each endpoint $P_{x}$, the error function can be noted as,
\begin{equation}
e_{k,P_{x}}^l=l \Pi( R_{k,j}P_{x}+t_{k,j}).
\label{e_l}
\end{equation}
%

To get a minimal parameterization of a plane $\pi$ for optimization, we represent it as $q(\pi) = (\phi, \psi, d)$ where $\phi$ and $\psi$ are the azimuth and elevation angles of the normal and $d$ is the distance from the Hessian form
\begin{equation}
q(\pi)=(\phi=arctan(\frac{n_y}{n_x}), \psi=arcsin(n_z), d).
\label{q}
\end{equation}
So, the error function between the observed plane $\pi_k$ in the frame and corresponding map plane $\pi_x$ is
\begin{equation}
e_{k,\pi_x}^{\pi}=q(\pi_k)-q(T_{cw}^{-T} \pi_x)
\label{e_plane}
\end{equation}
where $T_{cw}^{-T}$ is the transformation from world to camera coordinates.
%
Assuming that the observations follow a Gaussian distribution, the final non-linear least squares cost function $t*$ can be written as in~\eqref{eg:t*}, where  $\Lambda_{p_{k,j}}$, $\Lambda_{p_{k,P_x}}$ and $\Lambda_{k,\pi_x}$ are the inverse covariance matrices of points, lines and planes, and $\rho_p$, $\rho_l$ and $\rho_\pi$ are robust Huber cost functions, respectively.
\begin{equation}
\begin{aligned}
t* = &argmin\sum_{j}^M \rho_p\left({e_{k,j}^p}^T \Lambda_{p_{k,j}} e_{k,j}^p\right) \\
   & + \rho_l\left({e_{k,P_x}^l}^T\Lambda_{p_{k,P_x}} e_{k,P_x}^l\right) +\rho_\pi\left({e_{k,\pi_x}^{\pi}}^T\Lambda_{k,\pi_x} e_{k,\pi_x}^{\pi} \right)  \\
\end{aligned}
\label{eg:t*}
\end{equation}
Here, a solution is determined using the Levenberg-Marquardt algorithm.

\subsection{Pose refinement}
The last two steps assume that the scene is a good Manhattan model, nevertheless several general indoor environments are not strictly adhering to the MW assumption, leading to degradation in accuracy. So, after obtaining the initial pose via the decoupled rotation and translation strategy, the refinement module~\cite{Li2020SSLAM} fine-tunes the pose to compensate for deviations from the MW or unstable initial estimates. In the refinement step, to reduce the drift from frame-to-frame pose estimation, the local map constructed by previous keyframes is used to optimize the pose based on a map-to-frame strategy~\cite{murartal2017orb-slam2:}. 

Similar to~\cite{zhang2019point, murartal2017orb-slam2:,murartal2015orb-slam:}, we also use keyframes to build a local map, although our map has point, line and plane landmarks, which are projected into the current frame to search for matches. Furthermore, we explore the relationship between planes in the local map and planes detected in the current frame. The parallel and perpendicular constraints between those planes are described as~\eqref{e_parallel_plane},
\begin{equation}
\left\{\begin{array}{cc}
     e_{k,n_x}^{\pi_{\parallel}}=&||q_n(n_k)-q_n(R_{cw} n_x)||  \\
     e_{k,n_x}^{\pi_{\perp}}=&||q_n(R_{\perp} n_k)-q_n(R_{cw} n_x)||
\end{array}\right.
\label{e_parallel_plane}
\end{equation}
where $q_n(\pi)=(\phi, \psi)$ and $R_{cw}$ is the transformation from world to camera coordinates. For perpendicular planes, their plane normal is rotated by 90 degrees ($R_{\perp}$) to construct the error function.
These two error functions are merged to~\eqref{eg:t*} to build a joint optimization function in the refinement module.

\section{Mapping}
\label{sec:mapping}
This section describes the keyframe-based 3D mapping strategy used in our SLAM framework. Keyframes and 3D features build up a co-visibility graph, where nodes and edges are updated whenever a new keyframe and new features are available.
\subsection{Sparse Mapping}
\begin{figure}[t]
    \centering
    \includegraphics[scale=0.20]{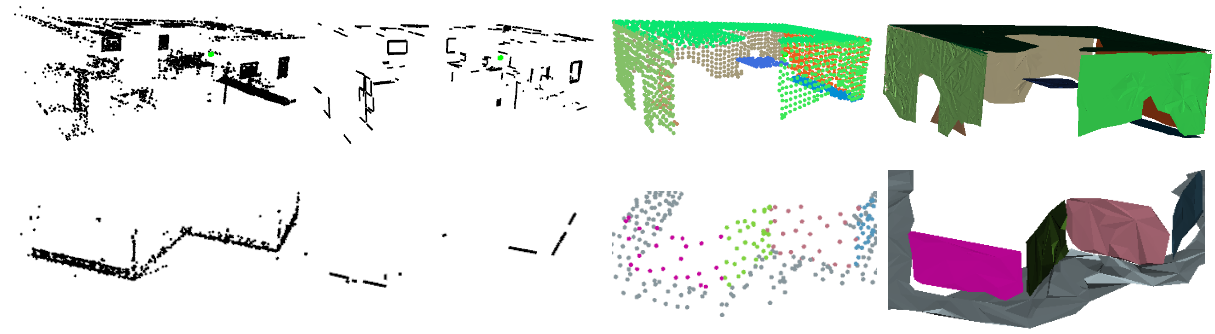}
    \caption{Different levels of maps provided by the system.Top row: office room of the ICL-NUIM; bottom row: structure-nontexture-near of TUM RGB-D; 
    }
    \label{fig:sparsemap}
\end{figure}

As shown in Figure~\ref{fig:sparsemap}, the sparse map module is reconstructed by point-line-plane features extracted from keyframes. The first frame is set as the first keyframe and the global map is initialized by the landmarks thereby detected. When new points, lines and planes are detected in a new keyframe, which are not in the global map, they will be saved to a local map first. Then we check the quality of the landmarks in the local map, and then push reliable landmarks into a global map after culling bad ones. Different to the matching methods for points and lines, for each detected plane in a new keyframe, we first check whether it is associated with a plane in the map using the strategy described in section~\ref{sec:structure_analysis}.
If we find an association, we add the 3D points of the new plane to the associated plane in the global map and filter out redundancies using a voxel grid to get a compact point cloud again. If the incoming plane is not associated to any plane in the global map, we add it to the map as a new plane.

\subsection{Planar instance-wise meshing}

The sparse map obtained in the previous section is still not adequate for applications involving robot-environment interactions, but it provides information about planar and non-planar instances. Therefore, we construct a denser map using an instance-wise meshing strategy.
Indoor scenes can be divided into planar and non-planar regions. Planar areas like floors, walls and ceiling have often a large extent, however a dense pixel-wise information does not add to the quality and is highly redundant. So instead of using surfel or TSDF, we regard plane regions as instances that include a small and fixed number of elements independently of their size.  

\begin{table*} [h]
\centering
\caption{Comparison of Translation RMSE (m) for ICL-NUIM  and TUM-RGB-D sequences. $\times$ means the method fails in the tracking process. -wo means only using decoupled tracking without the refinement step.}
\small
 \begin{tabular}{c|cccccccc}
\toprule
 Sequence & Ours &Ours/-wo & ORB~\cite{murartal2017orb-slam2:} &PS-SLAM~\cite{zhang2019point} &LPVO~\cite{kim2018linear} &L-SLAM~\cite{kim2018low} &DVO~\cite{kerl2013dense} & InfiniTAM~\cite{prisacariu2017infinitam}\\
 \hline
 lr-kt0 &\textbf{0.006} &0.025 &0.025  & 0.016  &0.015&0.012  &0.108 &$\times$ \\ 
 lr-kt1 &0.015&0.036 &0.008  & 0.018 &0.039&0.027 &0.059 &\textbf{0.006} \\ 
 lr-kt2&0.020  &0.053&0.023&0.017&0.034&0.053  &0.375 &\textbf{0.013} \\ 
 lr-kt3&\textbf{0.012} &0.059 & 0.021 & 0.025 &0.102&0.143 &0.433 &$\times$\\  \hline
 of-kt0 & 0.041 & 0.068& 0.037& 0.032 &0.061&\textbf{0.020} &0.244 &0.042 \\ 
 of-kt1 & 0.020&0.028 & 0.029& 0.019 &0.052&\textbf{0.015} &0.178 &0.025\\ 
 of-kt2 & \textbf{0.011} &0.060 & 0.039&0.026 &0.039&0.026  &0.099 &$\times$ \\
 of-kt3 & 0.014 &0.012 & 0.065&0.012 &0.030&0.011 &0.079 &\textbf{0.010}\\ \hline
 snot-far &0.022 &0.026& $\times$&\textbf{0.020} &0.075&0.141 &0.213&0.037\\
 snot-near & 0.025 & $\times$& $\times$& \textbf{0.013} &0.080&0.066 &0.076 &0.022\\
 cabinet &\textbf{0.035} &0.057 &0.075 &0.067 &0.520&0.291 &0.690 &\textbf{0.035}\\
 large-cabinet &\textbf{0.071} &0.813 & 0.124 &0.079  &0.279&0.140 &0.979&0.512\\
 \bottomrule
\end{tabular}
\label{table:ICL}
\end{table*}

In particular, we input plane instances to the meshing module, which meshes them independently. First, the points belonging to a plane are organized as a kd-tree data-structure. Different to unstructured inputs, our method needs less time for searching several nearest neighbors. Then, we use Greedy Surface Triangulation (GST)~\cite{Marton09ICRA} to build an instance-wise mesh, which is designed to deal with planar surfaces. Note that in our experiments, the initial search radius for selecting neighbors for triangulation is set to $5m$ and the multiplier is set as $5$ to modify the final search radius to adapt to different point densities on the plane regions.

\begin{figure}[t]
    \centering
    \includegraphics[scale =0.4]{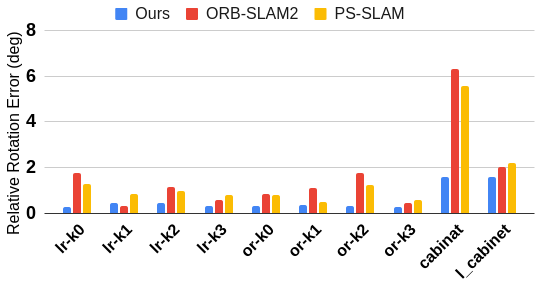}
    \caption{Comparison of relative pose error (RPE) for rotation on the ICL-NUIM and TUM RGB-D sequences.}
    \label{fig:rpe}
\end{figure}

\begin{figure}[t]
    \centering
    \includegraphics[scale=0.16]{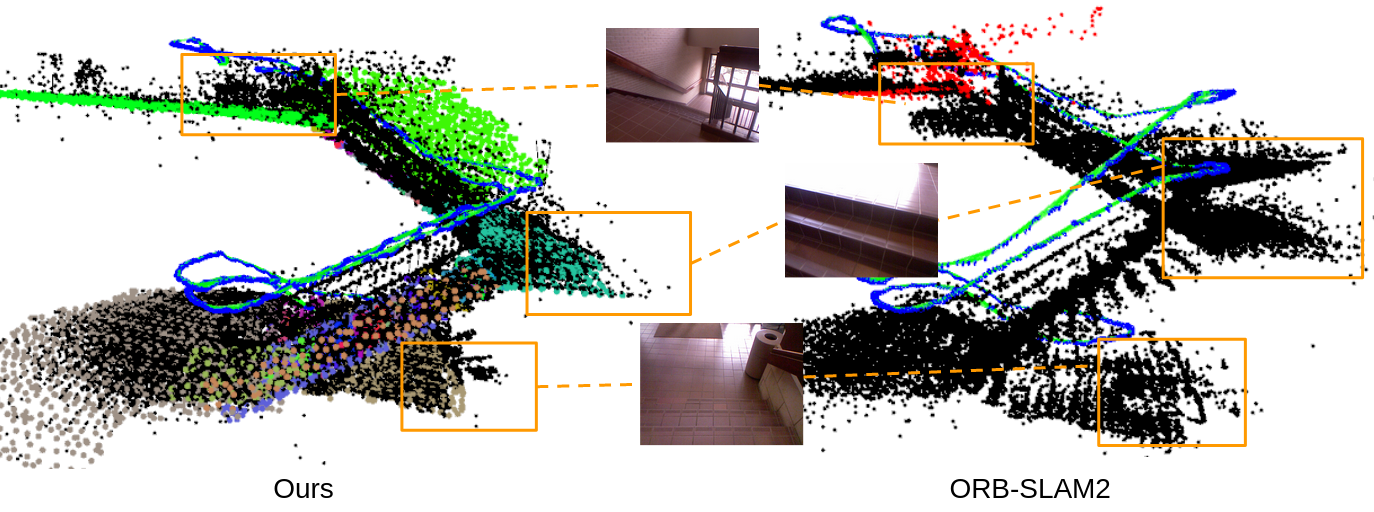}
    \caption{Qualitative results of sparse reconstruction and trajectory between the proposed method and ORB-SLAM2 in the TAMU dataset.}
    \label{fig:tamu}
\end{figure}
\section{Experiments}
We evaluate the proposed SLAM system on two well known public datasets, the ICL-NUIM~\cite{handa:etal:ICRA2014} and TUM RGB-D~\cite{sturm2012benchmark} benchmarks, comparing its performance with other state-of-the-art methods such as ORB-SLAM2~\cite{murartal2017orb-slam2:}, PS-SLAM~\cite{zhang2019point} that are feature-based methods, but removed the global bundle adjustment modules in the following experiments. Methods based on the MW assumption such as LPVO~\cite{kim2018low} and L-SLAM~\cite{kim2018linear}. DVO-SLAM~\cite{kerl2013dense} is a direct method and InfiniTAM~\cite{prisacariu2017infinitam} uses a GPU for real-time tracking and mapping based on RGB and depth images.
%
Additionally, we provide the reconstruction accuracy of our reconstructed model on the ICL-NUIM dataset and compare it with other popular methods for dense reconstruction. Lastly, to demonstrate that our system is robust over time, we also test on a sequence from the TAMU~\cite{lu2015robust} dataset containing long sequences covering a large indoor area. All experiments are carried out with an Intel Core i7-8700 CPU (with @3.20GHz) and without any use of GPU.
The ICL-NUIM dataset~\cite{handa:etal:ICRA2014} provides synthetic scenes for two indoor environments, one living room and one office room scenario. These scenes contain large areas of low textured surfaces such as walls, ceilings, floors, etc. There are four sequences for each scene. We evaluate our method on all sequences. 
\begin{table}[]
    \centering
    \caption{Measured tracking times (ms) on the TUM RGB-D sequences}
    \small
    \begin{tabular}{l|ccccc}
    \toprule
    time  & Feat. extr. &Rotat. &Transla & Refinement &Total \\    \hline
    Median  &19.9&2.1 &4.8 &13.0 &42.5  \\ 
    Mean  &20.5 &3.0 &5.4 &13.1 &43.7 \\ 
    Std. &3.6 &0.4 &2.8 &4.8&9.4\\ 
    \bottomrule
    \end{tabular}
    \label{tab:times}
\end{table}
\begin{table}[t]
\small
    \centering
    \small
    \caption{RMSE Reconstruction error (cm) on the ICL-NUIM dataset in centimeters.}
    \setlength{\tabcolsep}{1.5mm}{
    \begin{tabular}{c|ccccc}
       \hline
        Sequence &RGB-D &ElasticFu  &InfiniTAM    &SPFu  &Ours \\ \hline
        kt0   &4.4  &0.7 &1.3  &0.7 &\textbf{0.4}\\
        kt1   &3.2  &0.7 &1.1  &0.9          & \textbf{0.6}\\
        kt2   &3.1  &0.8  &\textbf{0.1} &1.1          &0.6\\
        kt3   &16.7 &2.8&2.8  &1.0 &\textbf{0.8}\\ \hline 
    \end{tabular}}
    \label{tab:recons_e}
\end{table}
\subsection{ICL-NUIM RGB-D Dataset}
Table~\ref{table:ICL} shows that our method obtains the best performance on three out of the eight sequences, based on the translation RMSE (ATE). InfiniTAM also performs well on lr-kt1, lr-kt2 and of-kt3 sequences, but the method also loses tracking in other sequences. As the dataset contains large structured areas, the Manhattan-based methods LPVO and L-SLAM are able to get a good estimate of the orientation and provide good results throughout. However, they usually need two planes, or alternatively, one plane and a vanishing direction to be visible at all times to estimate a good Manhattan frame. 
\begin{figure}[h]
    \centering
    \includegraphics[scale=0.29]{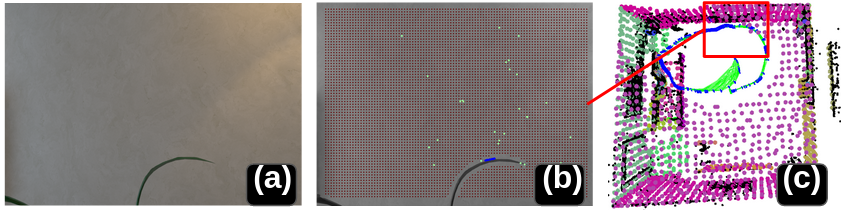}
    \caption{Results in lr-k3. (a) input image; (b) point-line features and the segmented plane; (c) reconstructed 3D map and trajectory.}
    \label{fig:lr-k3}
\end{figure}
As shown in Figure~\ref{fig:lr-k3}, there are several challenging scenes in lr-kt3, where only a white wall and two leaves from a plant are captured when the camera is close to the wall. In this situation, OPVO and L-SLAM are unable to yield a good performance. When a bad initial pose is obtained in our system due to the scene not being a rigid MW, the refinement step based on point-line-plane features allows us to recover the pose nevertheless, while L-SLAM ignores optimizing rotation in the LKF module. Moreover, while DVO, being a dense method, may struggle because of the large areas of walls, floor etc. not containing enough gradient for the photometric error, ORB-SLAM2 and PS-SLAM perform well, as both environments contain sufficient ORB features extracted from furniture, objects etc. As our method takes advantage of all geometric elements, it is able to perform robustly in most sequences. In addition, Figure~\ref{fig:rpe} shows the relative pose error for ORB-SLAM, PS-SLAM and our method. Our method obtains notably better results than the other two in relative translation and rotation. Especially the rotation error is much lower for our method, due to the use of the decoupled MW rotation estimation.
\begin{table}[t]
    \centering
    \small
     \caption{Comparison of the accumulated drift (m) in different large scale sequences.}
    \begin{tabular}{l|ccc}
    \toprule
  Sequences &Ours &ORB-SLAM2  &length \\ \hline
  Corridor-A  &1.62    &3.13  &88   \\
  Stair-A     &0.94   &1.44  &66   \\
  Entry-Hall &1.33    &2.22  &80   \\
  \bottomrule
    \end{tabular}
    \label{tab:my_label}
\end{table}
\subsection{TUM RGB-D Dataset}
The TUM RGB-D benchmark~\cite{sturm2012benchmark} is one of the most popular datasets for RGB-D SLAM systems, which provides indoor sequences under different texture and structure conditions. This allows us to separately test sequences which have structure, texture or both. In order to evaluate our method in challenging environments, we select four structured image sequences, the first three with low texture and the last one with a large scale environment. 
As all sequences listed in Table~\ref{table:ICL} have structure, but the large-cabinet sequence is not a rigid Manhattan scenario. Manhattan-based methods are able to provide good pose estimates on snot-far sequence, but the results degenerate in large-cabinet and cabinet sequences. The first two sequences include the same environment consisting of multiple non-textured planes. Here ORB-SLAM2 is not able to find enough point correspondences along the sequence and loses tracking. Our method, which additionally uses lines and planes for translation estimation, achieves better results. As shown in Figure~\ref{fig:rpe}, cabinet and large-cabinet are challenging sequences because of several low-texture frames. Our method's tracking strategy limits the relative rotation error to under 2 degrees, which is better than ORB-SLAM2 and PS-SLAM.
The statistics of the time spent for each operation are shown in Table~\ref{tab:times}, where we use different CPU threads to deal with points, lines and planes in the feature extraction and refinement modules. 

\subsection{Large scale sequence}
The TAMU dataset~\cite{Lu2015Robustness} provides large-scale indoor sequences (constant lighting). While it does not provide ground-truth camera poses, the start and end point are the same, which can be used to evaluate the overall drift by computing the final position errors.
%
As shown in Figure~\ref{fig:tamu}, the trajectory in the sequence Stair-C is a loop between two floors, where the improvement of our method over the whole trajectory length is 34.7\% in drift compared to ORB-SLAM2. Similar situations can also be found in Corridor-A and Entry-Hall. More qualitative results are provided in the supplementary material.

\subsection{Reconstruction Accuracy}
We reconstruct models from ICL-NUIM and compare the results with state-of-the-art mapping methods, as shown in Table~\ref{tab:recons_e}. The accuracy of the reconstruction results is defined as the mean difference between the predicted model and the ground-truth model~\cite{handa:etal:ICRA2014}. We compare the proposed mapping module against RGB-D SLAM~\cite{endres2012evaluation}, ElasticFusion~\cite{whelan2015elasticfusion}, InfiniTAM~\cite{kahler2016real}, and SuperpixelFusion~\cite{wang2019real}.

The SuperpixelFusion method is constrained by using ORB-SLAM for pose estimation, whereas our method also works well in low-textured environments. InfiniTAM obtains the best results in kt2, but shows worse performance on the kt0 and kt3 sequences, potential due to the large low-textured regions. ElasticFusion shows a similar behavior. Our method reconstructs more accurate maps than the others, but InfiniTAM and ElasticFusion provide more complete models than our map since we ignore small objects even though features based on points, lines and planes cover most of the pixels. Remarkably, all fusion methods, except for SuperpixelFusion and ours, rely on GPU based acceleration.

\section{Conclusions}
We have proposed a RGB-D SLAM system based on points, lines and planes. Using the MW assumption for rotation estimation, and point, line and plane features for translation estimation, we achieve state-of-the-art performance. Also, a novel instance-wise meshing approach can reconstruct planar regions in the environment efficiently. The resulting dense map allows for interactions with the environment in robotic and AR/VR applications.
In the future we would like to extend the planar reconstruction with a meshing of the non-planar parts in the environment to allow the complete reconstruction of more complex scenes.

{\small
\bibliographystyle{ieee}
\bibliography{egbib}
}

\end{document}